\documentclass[10pt,twocolumn,letterpaper]{article}

\usepackage{iccv}
\usepackage{times}
\usepackage{epsfig}
\usepackage{graphicx}
\usepackage{amsmath}
\usepackage{amssymb}
\usepackage{booktabs}
\usepackage{multirow}
\usepackage{mathtools}
\usepackage{subcaption}

\newcommand{\I}{\mathcal{I}}
\newcommand{\C}{\mathcal{C}}
\newcommand{\R}{\mathcal{R}}

\DeclarePairedDelimiter\floor{\lfloor}{\rfloor}

\graphicspath{{./images/}{./images/plots/}}
\DeclareGraphicsExtensions{.pdf}

\usepackage[pagebackref=true,breaklinks=true,letterpaper=true,colorlinks,bookmarks=false]{hyperref}

\usepackage{xcolor}

 \iccvfinalcopy 

\def\iccvPaperID{1677} 

\makeatletter
\def\blfootnote{\xdef\@thefnmark{}\@footnotetext}
\makeatother
 
\ificcvfinal\fi
\begin{document}

\title{LEWIS: Latent Embeddings for Word Images and their Semantics}

\author{Albert Gordo ~~~~~ Jon Almaz\'an ~~~~~ Naila Murray\\
Xerox Research Centre Europe\\
{\tt\small \{albert.gordo,jon.almazan,naila.murray\}@xrce.xerox.com}
\and
Florent Perronnin\\
Facebook AI Research $^*$\\
{\tt\small perronnin@fb.com}
}

\maketitle

\blfootnote{$^*$ This work was done while Florent Perronnin was affiliated with the Computer Vision Group at Xerox Research Centre Europe.}

\begin{abstract}
The goal of this work is to bring semantics into the tasks of text recognition and retrieval in natural images.
Although text recognition and retrieval have received a lot of attention in recent years, previous works have focused on recognizing or retrieving \emph{exactly} the same word used as a query, without taking the semantics into consideration. 

In this paper, we ask the following question: \emph{can we predict semantic concepts directly from a word image, without explicitly trying to transcribe the word image or its characters at any point?}
For this goal we propose a convolutional neural network (CNN) with a weighted ranking loss objective that ensures that the concepts relevant to the query image are ranked ahead of those that are not relevant.
This can also be interpreted as learning a Euclidean space where word images and concepts are jointly embedded.
This model is learned in an end-to-end manner, from image pixels to semantic concepts, using a dataset of synthetically generated word images and concepts mined from a lexical database (WordNet).
Our results show that, despite the complexity of the task, word images and concepts can indeed be associated with a high degree of accuracy.
\end{abstract}

\section{Introduction}

In recent years there has been an increased interest in tasks related to {\em text recognition and retrieval} 
in natural images \cite{Karatzas:2013,Wang:2011}.
For example, given an image of a word, one may be interested in recognizing the word, 
either using a list of possible transcriptions \cite{Almazan:2014,Gordo:2015,Wang:2011} 
or in an unconstrained manner \cite{Bissacco:2013,Jaderberg:2015}.
There has also been a growing interest in word image retrieval: 
given a query, which can be either a text string or another word image, 
one tries to retrieve the relevant word images in a dataset \cite{Almazan:2014,Gordo:2015}.

\begin{figure}[t!]
\centering
\includegraphics[width=0.99\linewidth]{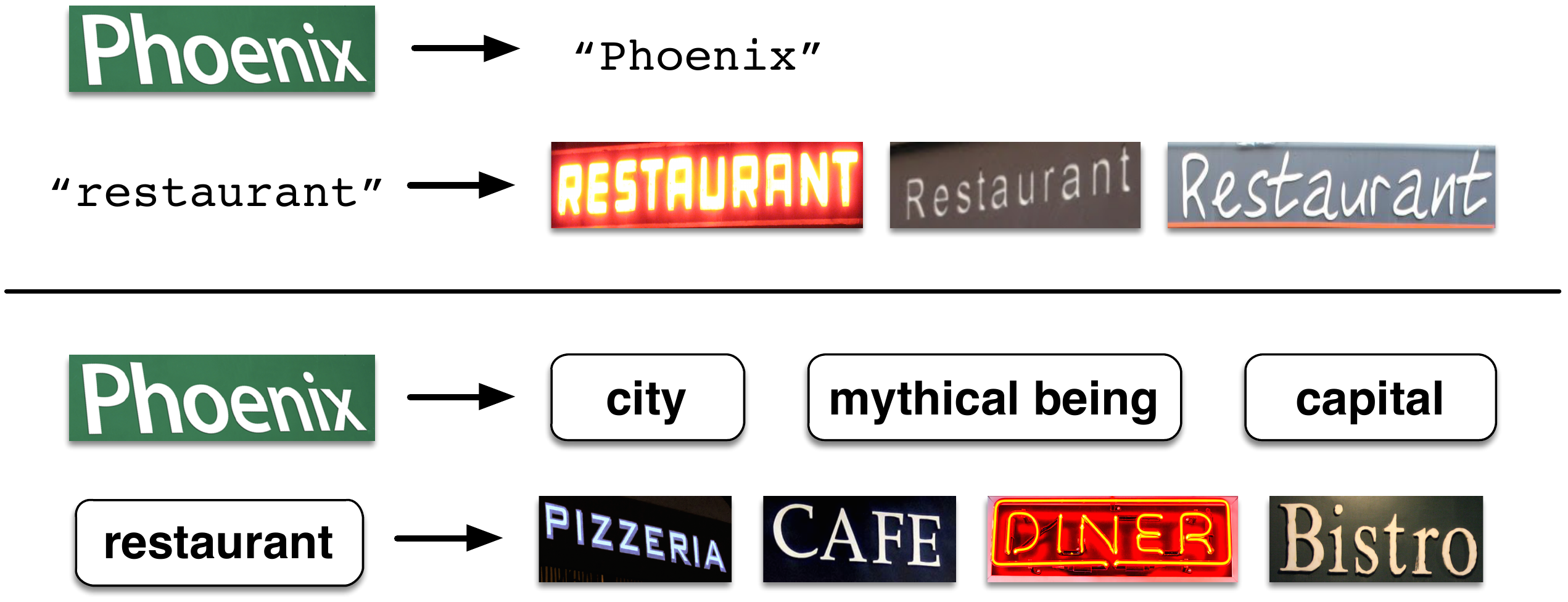}
\caption{Comparison of standard scene text recognition and retrieval (top), and the proposed word image understanding tasks (bottom). Strings in quotes represent text strings while strings in bounding boxes represent concepts.}
\label{fig:intro}
\vspace{-0.4cm}
\end{figure}

In all these cases, the goal has been to retrieve or recognize \emph{exactly} 
the same word used as a query, without taking the semantics into consideration. 
For example, given a query image with the word \texttt{phoenix}, it would be transcribed as \texttt{phoenix}, without any consideration of its meaning.
Similarly, using the text string \texttt{restaurant} as a query would only retrieve images containing this word in them (see Figure~\ref{fig:intro} top).

In contrast, in this paper we are interested in the problem of \emph{word image understanding},
\ie we wish to bring \emph{semantics} into the tasks of word image recognition and retrieval.
For example, we would like to capture the semantic meanings of the word \texttt{phoenix} as both a city and a state capital, and also its semantic meaning as a mythical being (see Figure~\ref{fig:intro} bottom).
Semantics play a very important role in scene understanding and for scene text, particularly in urban scenarios,
they will allow one to perform tasks beyond simple lexical matching.
To illustrate this, let us take the example of a system which would parse a street scene and especially
which would classify building faces into different business classes such as restaurants, hotels, banks, etc.
While the presence of a sign \texttt{pizzeria} is indicative of a restaurant,
the mere transcription of the text in the sign is not sufficient in itself to deduce this.
Additional reasoning capabilities enabled by an understanding of the semantics of the word are required to make the classification decision.

A straightforward, two-step approach to achieving this goal would be to first transcribe the word image, 
and then match the transcriptions to the semantic concepts. 
The transcriptions could be matched using lexical databases of English such as WordNet \cite{wordnet}, 
that contain a  hierarchy of words annotated with semantic concepts.
In this two-step approach, the word-image recognition  step can be understood as 
one of extracting mid-level features -- the transcriptions -- which are then fed to a second classification step.

However, this approach has significant shortcomings.
First, it relies on an accurate transcription of word images.
Although the state-of-the-art in word image recognition 
has significantly leaped forward in recent years~\cite{Almazan:2014,Gordo:2015,Jaderberg:2014b,Jaderberg:2015}, 
the results are still not perfect, particularly when word images are not cropped exactly.
This is a very typical scenario in end-to-end word recognition, 
where one first has to localize the word in the image, crop it, and then recognize it.
Second, the approach cannot deal with out-of-vocabulary words. 
Even if a word is transcribed correctly, if the word does not appear in the lexical resource, it will not be possible to assign concepts to it. 
Finally, this approach does not lead to a compact representation of word images that encodes semantics.
Such a representation is desirable as it could be used as an input feature for other tasks such as clustering word images that share semantics, or
searching among word images using a semantic concept as a query -- see Figure~\ref{fig:qres_c2i} for an example.

\begin{figure}[]
\centering
\includegraphics[width=0.99\linewidth]{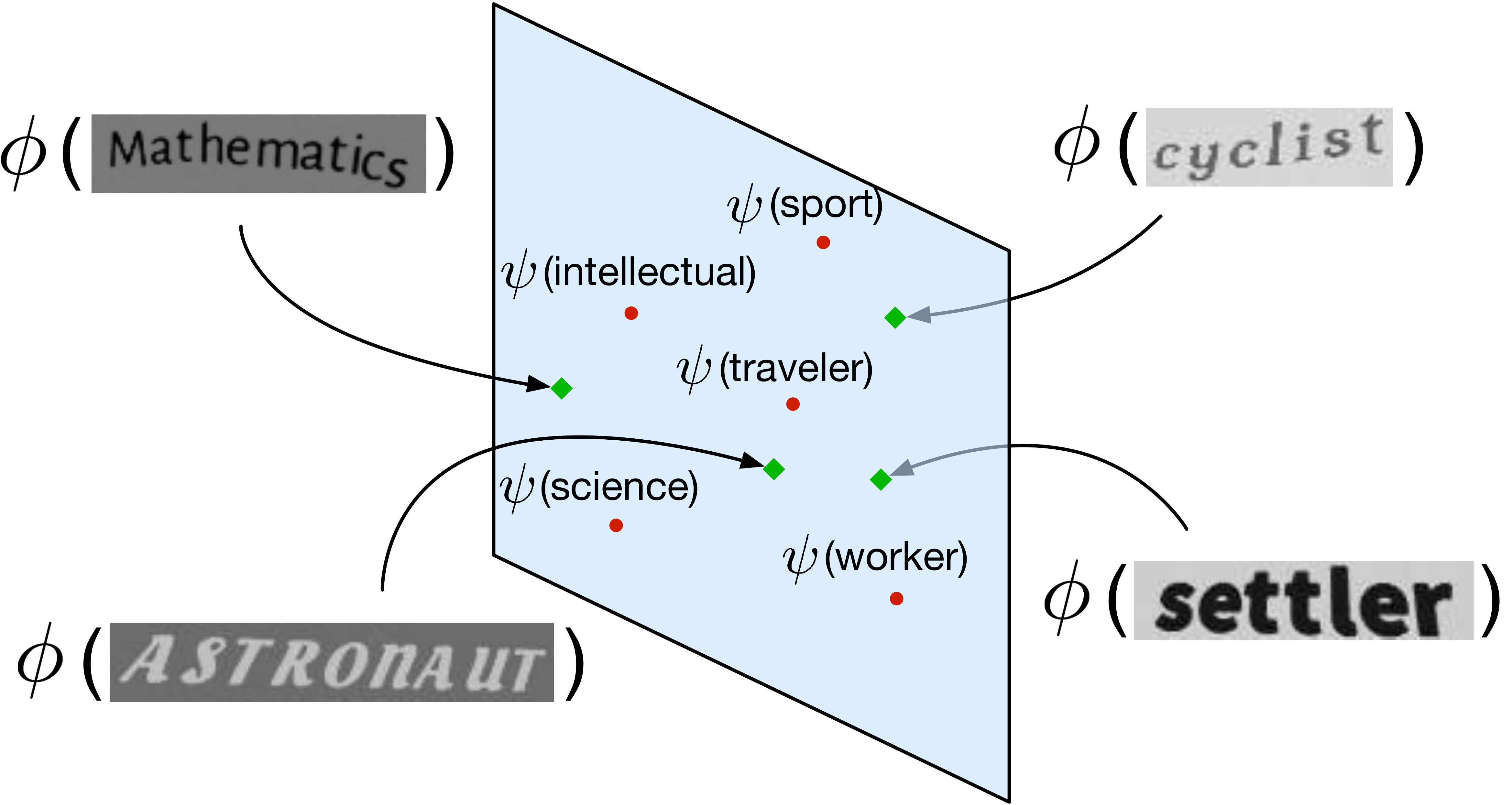}
\caption{Outline of our approach.
Our goal is to learn two embedding functions $\phi:~\I~\rightarrow~\R^D$ and $\psi:~\C~\rightarrow~\R^D$ that embed images and concepts in a common subspace, 
and where embedded images should be closer to the embedded concepts they are related with than to the concepts they are not related with.
This is learned in an end-to-end manner with a convolutional neural network and a ranking loss.}
\label{fig:diagram}
\end{figure}

In this paper, we ask the following question: \emph{can we predict semantic concepts directly from a word image
without explicitly trying to transcribe the word image or its characters at any point?}
While this might sound hopeless, because different word images corresponding
to the same concept may have widely varying appearances (see the \texttt{restaurant} example in Figure~\ref{fig:intro} bottom),
we show that, surprisingly, this is indeed possible.
For this goal we propose to use a convolutional neural network (CNN) \cite{Lecun:1989} 
with a weighted ranking loss objective~\cite{Weston:2011} that ensures that the concepts relevant 
to the query image are ranked ahead of those that are not relevant.
This model is learned in an end-to-end manner, 
from image pixels to semantic concepts.
Importantly, one can interpret this learned architecture as a way to embed word images and concepts in a common, latent subspace (see Figure~\ref{fig:diagram}).
In particular, the weights of the last layer of the network can be seen as a \emph{transductive embedding} of the semantic concepts 
-- to add new concepts, one would need to retrain or fine-tune the network.
On the other hand, the activations of the previous-to-last layer of the network can be seen as an \emph{inductive embedding} of the input word images: 
word images containing words that have not been observed during training can still be embedded in this space and matched with known concepts.
Hence, we refer to our approach as Latent Embeddings for Word Images and their Semantics, or LEWIS for short.

LEWIS addresses the problems of the straightforward two-step approach: 
it does not require one to extract the transcription of word images explicitly, 
it allows one to retrieve concepts from word images not seen during training or that do not appear in our source of semantic concepts 
-- akin to a zero-shot learning task -- 
and it provides a feature vector representation of the word images that encodes its semantics rather than only its lexical information.
As we will show in later sections, this allows one to go {\em beyond} predicting semantic categories from word images and perform additional tasks such as 
searching word images using a concept as a query, or retrieving word images that share concepts with a query word image, 
even when both images depict different words.

In summary, the contributions of the paper are four-fold.
First, we introduce a new task to the computer vision community:
predicting semantic categories of word images.
Second, we enrich an existing large dataset of word images with semantic transcriptions derived from WordNet \cite{wordnet} to evaluate the proposed problem -- 
we plan to make these annotations available to the community.
Third, to solve the proposed problem we introduce LEWIS, a solution based on a convolutional architecture which does not involve
transcribing the word image and which embeds both word images and semantic concepts in a latent common subspace.
Fourth, we show experimentally that LEWIS performs comparably or better in terms of accuracy than a two-step approach 
that uses state-of-the-art word recognition techniques, while offering many other advantages.

The rest of the paper is organized as follows.
Section~\ref{sec:related} describes the related work.
Section~\ref{sec:embeddings} introduces our method.
In Section~\ref{sec:experiments} we describe our experimental evaluation and discuss the results.
Finally, Section~\ref{sec:conclusions} concludes the paper.

\section{Related Work}
\label{sec:related}

We review the most related works to ours:
those related to word image representations (embeddings), 
textual embeddings, and
joint image and semantic embeddings.

\paragraph{Word image representations.}
Deriving suitable representations of word images for tasks such as recognition and retrieval in document images 
has been an active topic of research in the document analysis community for many years.
However, only during recent years has that interest also embraced word image representations in natural images, 
commonly referred to as ``scene text'' \cite{Almazan:2014, Bissacco:2013, Gordo:2015, Jaderberg:2014a, Jaderberg:2014b, Mishra:2012a, Mishra:2012b, Neumann:2013,  Rodriguez:2014, Rodriguez:2013,Shi:2013, Yao:2014}.

Many of these works focus on localizing the individual characters of the word image. 
Then, one may recognize the characters independently to produce a transcription \cite{Bissacco:2013,Neumann:2013}, 
or define a compatibility function between the character probabilities and text strings (using \eg conditional random fields) 
and rank all possible words in a text dictionary or lexicon to find the most likely transcriptions \cite{Jaderberg:2014a, Mishra:2012a, Mishra:2012b, Shi:2013, Yao:2014}.
The previous approaches do not explicitly construct a feature representation of the word image, and their uses beyond recognition are limited. 
In contrast, some recent works \cite{Almazan:2014, Gordo:2015, Rodriguez:2014, Rodriguez:2013} 
focus on obtaining a feature representation of the image using standard computer vision representations, 
without explicitly localizing its characters, and learning a compatibility function between these feature vector representations and embedded text strings. 

On a slightly different line, and closely related to our approach, 
Jaderberg \etal~\cite{Jaderberg:2014b} learn to classify words images into a set of $90{,}000$ possible transcriptions. 
This is achieved in an end-to-end manner using Convolutional Neural Networks (CNNs) and synthetic training data, 
and the approach obtained outstanding recognition results on standard benchmarks.
Interestingly, the activations of the previous-to-last layer of the network can also be used as word image features for retrieval purposes. 

All these previous works focus solely on lexical similarities, 
and do not capture any semantics.
On the contrary, we focus on capturing the semantic information of the words, 
and not simply information about the word transcription.

The most closely related work to ours is the one by Krishnan and Jawahar~\cite{Krishnan:2013}, 
that aims at performing word image retrieval preserving semantics, and, particularly, synonyms. 
This is very similar to the two-step baseline discussed in the previous section.
Contrary to us, \cite{Krishnan:2013} does not learn any joint space for images and semantics, and relies on query expansion using a dataset of images annotated with their synonyms. 
\vspace{-0.15cm}
\paragraph{Text embeddings.}
There has been a recent resurgence of interest in embedding text in semantic Euclidean spaces in the natural language processing community. 
Examples of such works include Word2Vec \cite{Mikolov:2013} and GloVe \cite{Pennington:2014}. 
This is achieved by unsupervised training on large corpora of text such as Wikipedia.
However, these approaches focus on embedding text strings, and not word images.

\paragraph{Images and their semantics}
Several works have considered the problem of jointly embedding images and semantic categories 
in an intermediate Euclidean space.
A simple way to do so is to perform a Canonical Correlation Analysis (CCA)
on image representations and their tags~\cite{Gong:2013}.
Weston \etal~\cite{Weston:2011} proposed WSABIE which can be understood
as a neural architecture with a single hidden layer.
The WSABIE objective function is a weighted ranking loss.
We use a similar loss to learn our joint word-image and semantic concept embedding.
An issue with WSABIE is that it cannot deal with zero-shot recognition.
To address this problem, Frome \etal~\cite{Frome:2013} proposed Devise, 
an embedding model that learns to map natural images to text embeddings learned with Word2Vec.
Other recent works also used text embeddings as output embeddings \cite{Akata:2014,Akata:2015}, but focusing only on natural images.
By contrast, we do not leverage the Word2Vec representations of text, 
and rely on the graph taxonomy provided by WordNet \cite{wordnet} to learn our embeddings.

\section{Learning latent embeddings}
\label{sec:embeddings}
We start by describing how we mine WordNet for semantic concepts.
Then we describe our approach to ranking those concepts given an image, and how it can be understood as an embedding of word images and concepts in a common latent subspace.

\subsection{Mining WordNet for Semantic Concepts}
\label{sec:wordnet}
\begin{figure}[]
\centering
\includegraphics[width=0.95\linewidth]{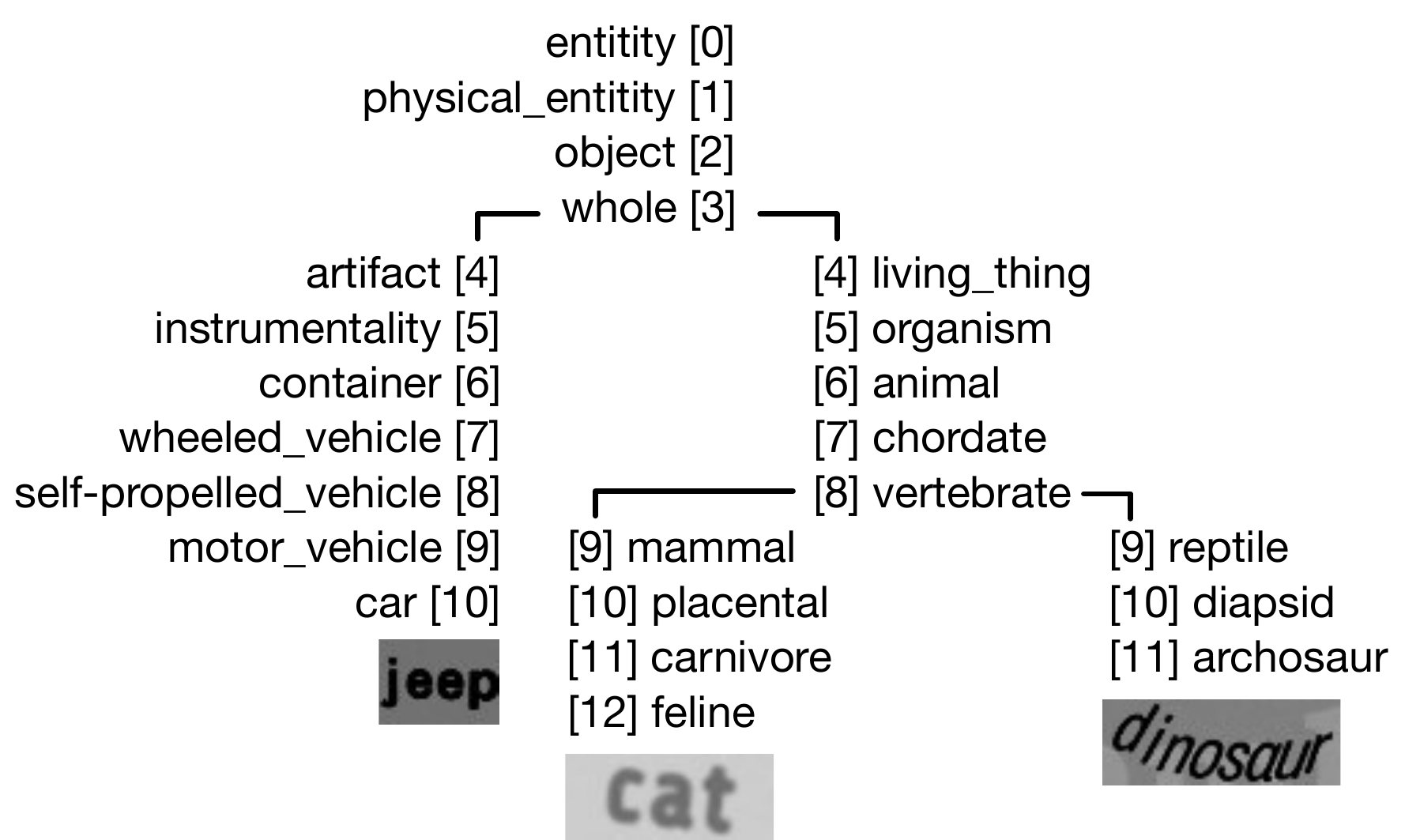}
\caption{A section of the WordNet hierarchy showing three words and their hypernyms up to the root. Note that \texttt{cat} and \texttt{dinosaur} would be given the same label for depth level $8$ and above, but different labels otherwise. On the other hand, \texttt{jeep} and \texttt{dinosaur} would not share concepts until reaching depth level $3$.}
\label{fig:wn_tree}
\end{figure}

WordNet \cite{wordnet} is a lexical database for the English language.
Words are organized into groups of synonyms called \emph{synsets}, and these groups in turn are organized in a hierarchical manner using different semantic relations.
One of these types of relations is hypernymy: Y is an hypernym of X if X is a kind of Y.
For example, the word \text{glass} has several hypernyms, two of which are \texttt{solid} (when \texttt{glass} is a material) and \texttt{container}. 
Therefore, given a word, one can find the synset or synsets (if the word has several meanings) to which it belongs, and then \emph{climb} through the hypernym hierarchy until a root is found.
As an example, paths for the words \texttt{jeep}, \texttt{cat}, and \texttt{dinosaur} are shown in Figure~\ref{fig:wn_tree}, where the number within brackets indicates the depth level of the hierarchy.

In our work, we leverage these hierarchies to produce semantic annotations of words: given a word in our dataset, we first produce the set of synsets to which it belongs. Each synset in this set corresponds to a different, fine-grained, semantic meaning of the word. Then, for each synset in the set, we ascend the hypernym hierarchy, producing increasingly generic concepts with which we annotate the word.
Annotating words with all of their hypernyms would produce tens of thousands of concepts, some of them very fine grained (\eg \texttt{goblet}), while some others being extremely generic (\eg \texttt{entity}).
Instead, we collect only the concepts at a given depth level, controlling the granularity of the concepts.
For example, when choosing level $9$, \texttt{cat} would be labeled as \texttt{mammal} and \texttt{dinosaur} as \texttt{reptile}, while at level $8$ both would be labeled as \texttt{vertebrate}.
We evaluate and compare the results for different choices of the depth level.

This annotation approach still produces several thousands of classes, some of which are very populated while others contain as few as one single word.
For evaluation purposes, we will only annotate words with the $K$ most populated classes and evaluate the effect of changing the value of $K$.

It is worth noting that, although we base our evaluation on semantic concepts extracted from WordNet, this is not a requirement, and other sources of semantic annotations can be exploited by our method.

\subsection{Ranking semantic concepts}

Motivated by the overwhelming success of convolutional neural networks (CNN) \cite{Lecun:1989} 
for image classification  \cite{Krizhevsky:2012} and word image recognition \cite{Jaderberg:2014b}, 
we adopt a similar architecture.
We follow Jaderberg et al \cite{Jaderberg:2014b} and use
5 convolutional layers followed by $3$ fully connected layers.
In our case the output dimensionality of the last fully connected layer is the number of semantic concepts $K$.
Please refer to Section \ref{sec:implementation} for more details about the network architecture.

A significant difference with  \cite{Krizhevsky:2012,Jaderberg:2014b} is that 
these works address a mono-label classification problem
while we consider a multi-label problem since multiple concepts can be assigned to an image.
Hence, we cannot adopt the standard classification objective which involves computing
the cross-entropy between the network output and the ground truth label.
Instead, we make use of a ranking framework which we now explain.

Let us assume a set of $N$ training images $\{\I_1,\I_2,\ldots,\I_N\}$, 
and a set of $K$ semantic concepts $\{\C_1,\C_2,\ldots,\C_K\}$ produced as described in the previous section.
Let us also assume that, for training purposes, each image is annotated with at least one semantic concept.
As the transcriptions of the word images are available at training time, 
this can be achieved by propagating to each image the concepts that are relevant to its transcription. 
Let us denote by $r(\I)$ the set of concept indexes that are relevant to image $\I$ 
and by $\bar{r}(\I)$ its complementary.

Given this setup, we are interested in finding a compatibility function $F$ between images 
and concepts such that the number of non-relevant concepts that are ranked ahead of relevant concepts is minimized.
Given an image $\I$, the last fully connected layer of the architecture produces a prediction vector $Y \in \R^K$, 
where $Y_i$ represents the predicted compatibility between the image and concept $\C_i$, \ie $Y_i = F(\I, \C_i)$.
A possible ranking objective is to enforce: 
\begin{equation}
    \min_F \sum_{\I} \sum_{\substack{p \in r(\I)\\ n \in \bar{r}(\I)}} \mathbf{I}_{[ F(\I, \C_n) > F(\I, \C_p)]}, 
    \label{eq:rank}
\end{equation}
where $\mathbf{I}_{[cond]}$ is the indicator function that evaluates to $1$ when $cond$ is true and to $0$ otherwise.
However, optimizing Equation \eqref{eq:rank} directly is not feasible due to the indicator function, and instead we choose a differentiable surrogate.
In particular, we choose the weighted approximately ranked pairwise loss (WARP) of Weston \etal~\cite{Weston:2011}.
This ranking loss places more emphasis on the top of the ranked list, leading to superior results under many ranking metrics.

Given two concepts $p \in r(\I)$ and $n \in \bar{r}(\I)$, their WARP loss is computed as
\begin{equation}
    \ell(\I,p,n) = L(rank(p)) \cdot max(0, 1 - Y_p + Y_n).
\end{equation}

Here, $rank(p)$ denotes the ranked position of $p$, \ie, how many concepts obtained a better score than $\C_p$, while $L(r)$ is a loss function of the form:
\begin{equation}
    L(r) = \sum_{j=1}^r \alpha_j,\; \text{with}\; \alpha_1 \geq \alpha_2 \geq \ldots \geq 0,
\end{equation}
where different choices of the $\alpha_j$ coefficients lead to the optimization of different measures, and where $\alpha_j = 1/j$ puts special emphasis on the first results, leading to superior top $K$ accuracy and mean average precision \cite{Weston:2011}.

Computing the loss over all possible pairs of $p$ and $n$ may be prohibitively expensive.
Instead, given an image and a positive category (or concept in our case), one typically samples negative categories until finding one which produces a positive loss, and uses that for the update.
Similarly, computing the exact rank of $p$ is expensive if $K$ is not small.
In that case, the rank of $p$ can be estimated as $\floor{\frac{K-1}{s}}$, where $s$ is the number of tries that was needed to find a negative category with a positive loss.
Although this approximation is rough, particularly for items with multiple positive labels, it works well in practice \cite{Weston:2011}.

The subgradient of the loss, needed for the backpropagation stage of the training, is given by:
\begin{equation}
    \frac{\partial \ell(\I,p,n)}{\partial Y_i} = 
     \begin{cases} 
         -L(rank(p))  & i= p \mbox{ and } \ell(\I,p,n) > 0, \\
         L(rank(p))   & i= n \mbox{ and } \ell(\I,p,n) > 0, \\
         0 & \text{otherwise}.
               \end{cases}           
\end{equation}

\subsection{Latent embeddings}
It has been shown in several recent works that CNNs can be used as generic feature extractors, 
and that these features are useful for tasks such as classification and retrieval \cite{Chatfield:2014, Babenko:2014}.
This is achieved by using the output activations of a given layer of the network.
For example, the output activations of the last layer produces a task-specific  ``attributes'' representation that encodes 
the scores that the image obtains for each of the classes used during learning.
Extracting features from earlier layers produces more and more generic features, 
which are more and more disconnected from the learning objective~\cite{Yosinsky:2014}.

Here we follow a similar idea and use the activations of the penultimate layer of our architecture (FC7, $4{,}096$ dimensions) 
as semantic representations of the word images.
However, we also note that the columns of the weight matrix of the last layer can be seen as embeddings of the semantic concepts.
What is more, because of the way the network is constructed, the dot product between the word embeddings and the concept embeddings is exactly the compatibility function $F$ between word images and concepts that we were seeking.
If we denote by $\phi(\I)$ the activations of the FC7 layer of the network given image $\I$, 
and by $\psi_k$ the $k$-th  column of the weight matrix of the last layer, 
then $F(\I, \C_k) = \phi(\I)^T \psi_k$.
$\psi_k$ can be understood as a transductive embedding of concept $\C_k$, 
\ie we can define a function $\psi$ which acts as a simple look-up table such that $\psi(\C_k) = \psi_k$.
This interpretation gives better insights into some of the tasks we perform such as querying the image dataset using a concept as a query, 
or performing an image-to-image search, as illustrated in Figure~\ref{fig:diagram}.

\vspace{-0.1cm}
\section{Experiments}
\label{sec:experiments}
We start by describing our datasets.
We then describe our evaluation protocols and baselines and provide quantitative as well as qualitative results.

\begin{figure*}[t!p]
        \captionsetup[subfigure]{justification=centering}
         \begin{subfigure}[b]{0.24\textwidth}
                 \centering
                 \includegraphics[width=0.85\textwidth]{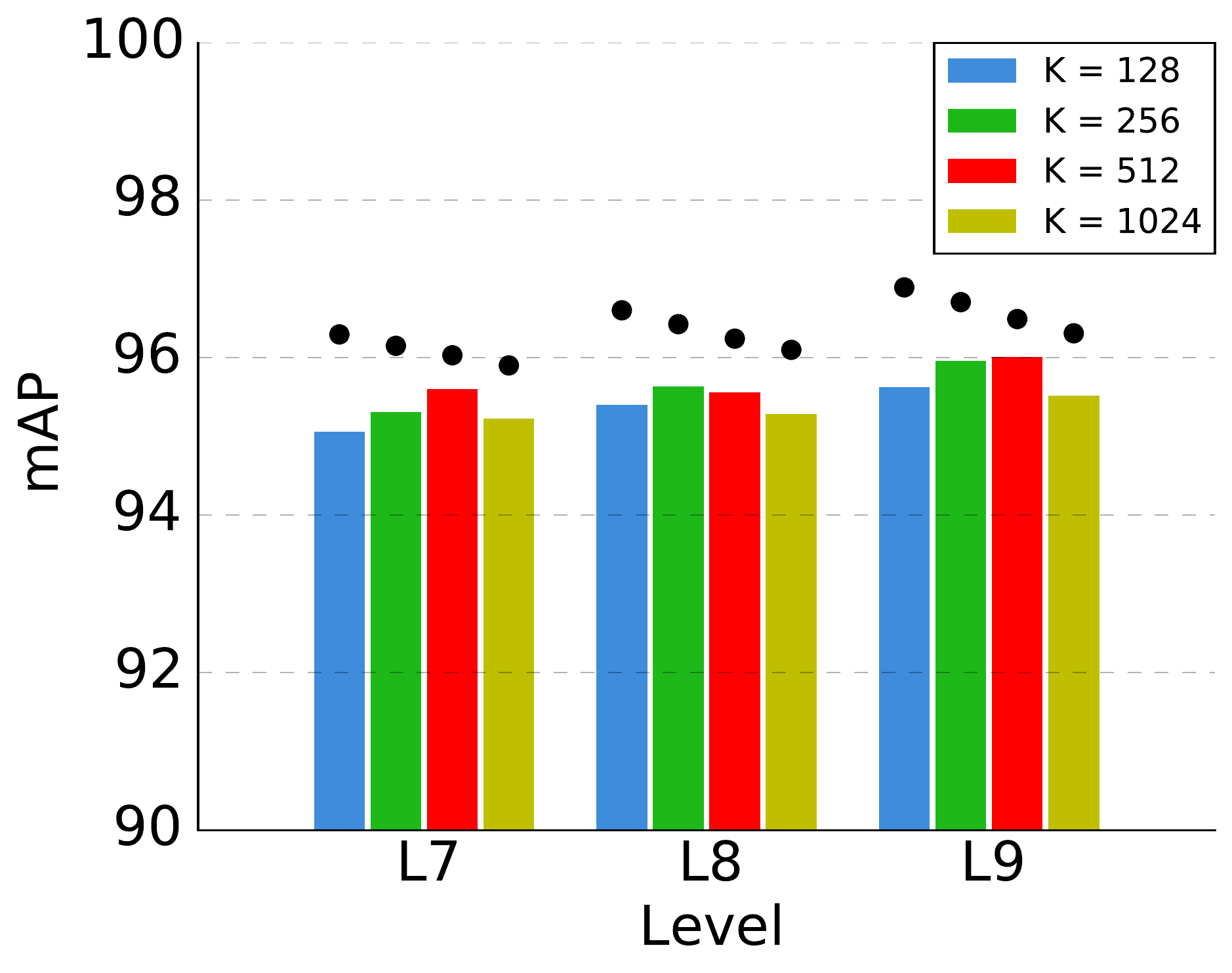}
                 \caption{\small image-to-concept \\(accurate localization)}
                 \label{fig2:i2c}
         \end{subfigure}
         \begin{subfigure}[b]{0.24\textwidth}
                 \centering
                 \includegraphics[width=0.85\textwidth]{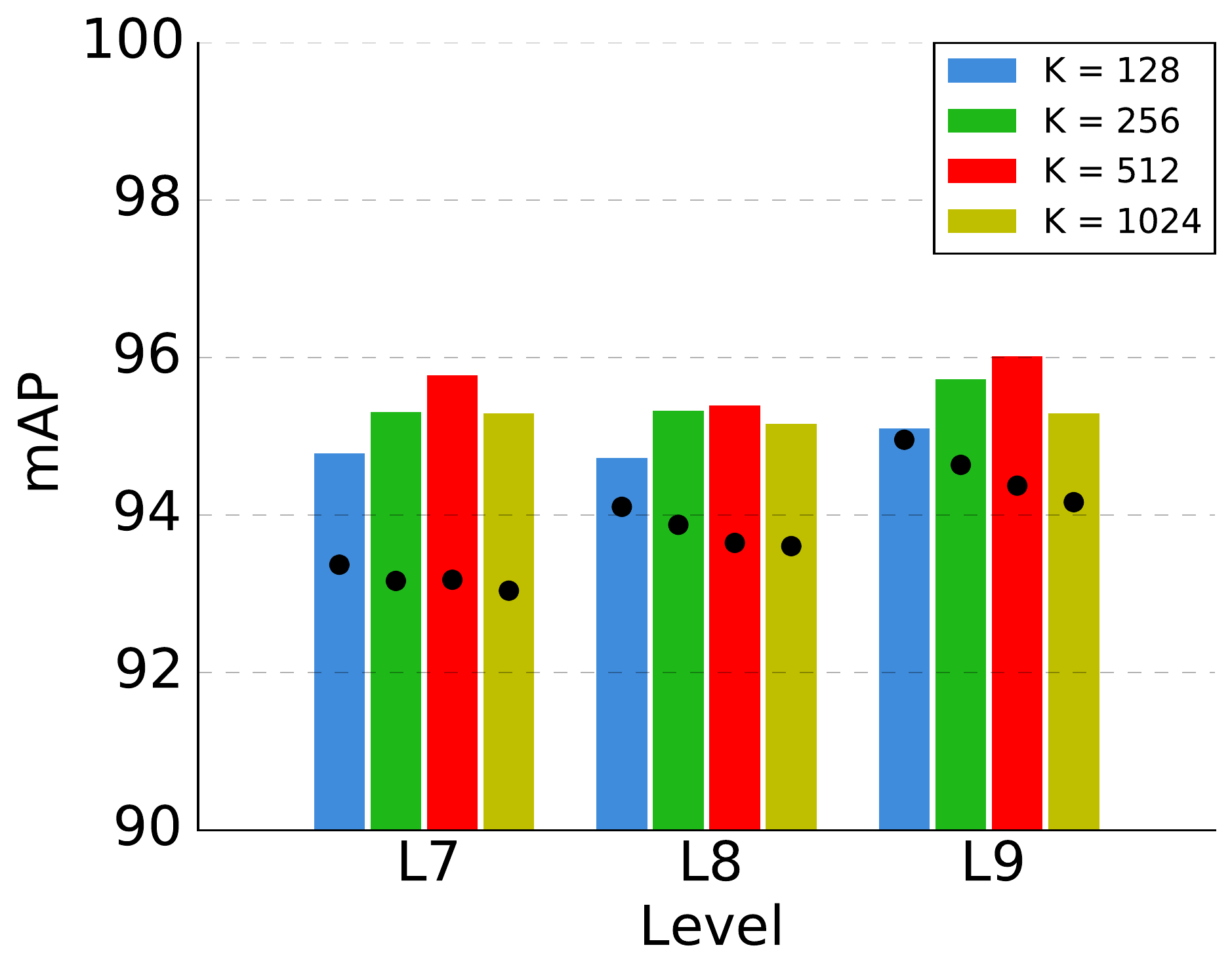}
                 \caption{\small concept-to-image \\(accurate localization)}
                 \label{fig2:c2i}
         \end{subfigure}%
         \begin{subfigure}[b]{0.24\textwidth}
                 \centering
                 \includegraphics[width=0.85\textwidth]{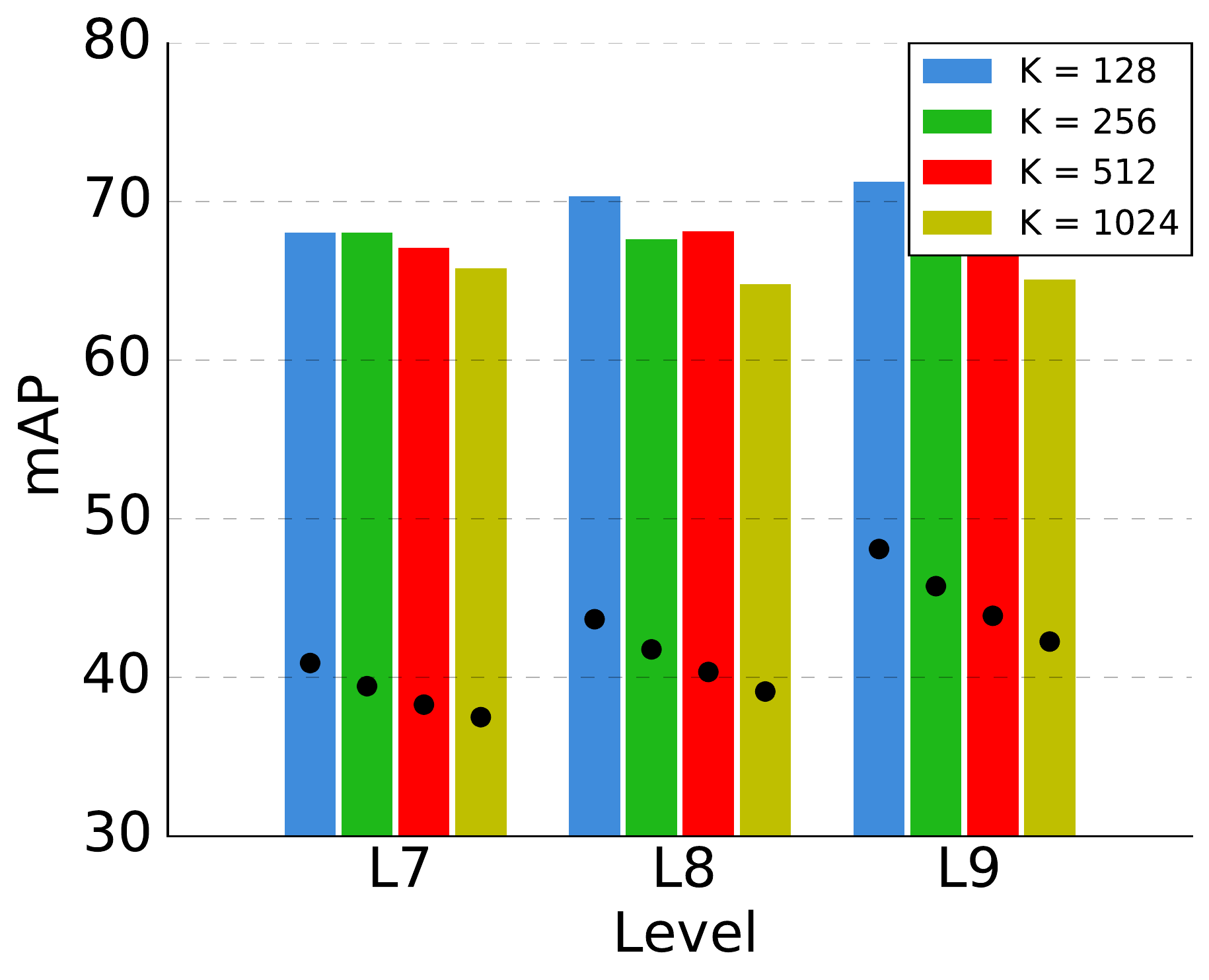}
                 \caption{\small image-to-concept \\(inaccurate localization)}
                 \label{fig2:i2cr}
         \end{subfigure}
         \begin{subfigure}[b]{0.24\textwidth}
                 \centering
                 \includegraphics[width=0.85\textwidth]{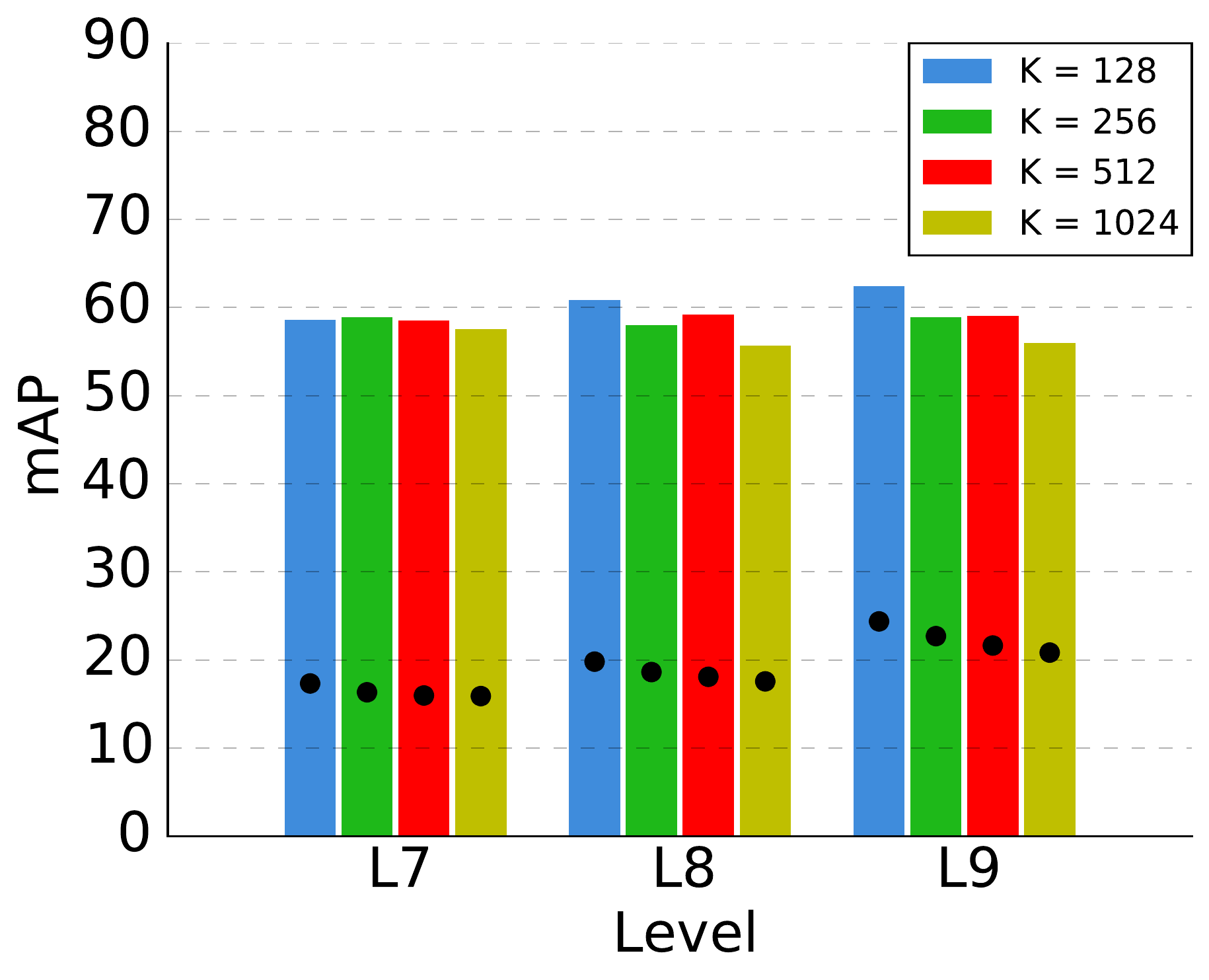}
                 \caption{\small concept-to-image \\(inaccurate localization)}
                 \label{fig2:c2ir}
         \end{subfigure}
        \caption{Quantitative results on Oxford Synthetic. 
          The bars represent the accuracy of our proposed LEWIS method, while the dots represent the accuracy of the two-step baseline.
        (a), (b): image-to-concept and concept-to-image results on the original word images, which were accurately cropped. 
          (c), (d): results on random crops of the word images.}
        \label{fig:results_oxford}
\end{figure*}

\begin{figure*}[t!p]
        \captionsetup[subfigure]{justification=centering}
         \begin{subfigure}[b]{0.24\textwidth}
                 \centering
                 \includegraphics[width=0.85\textwidth]{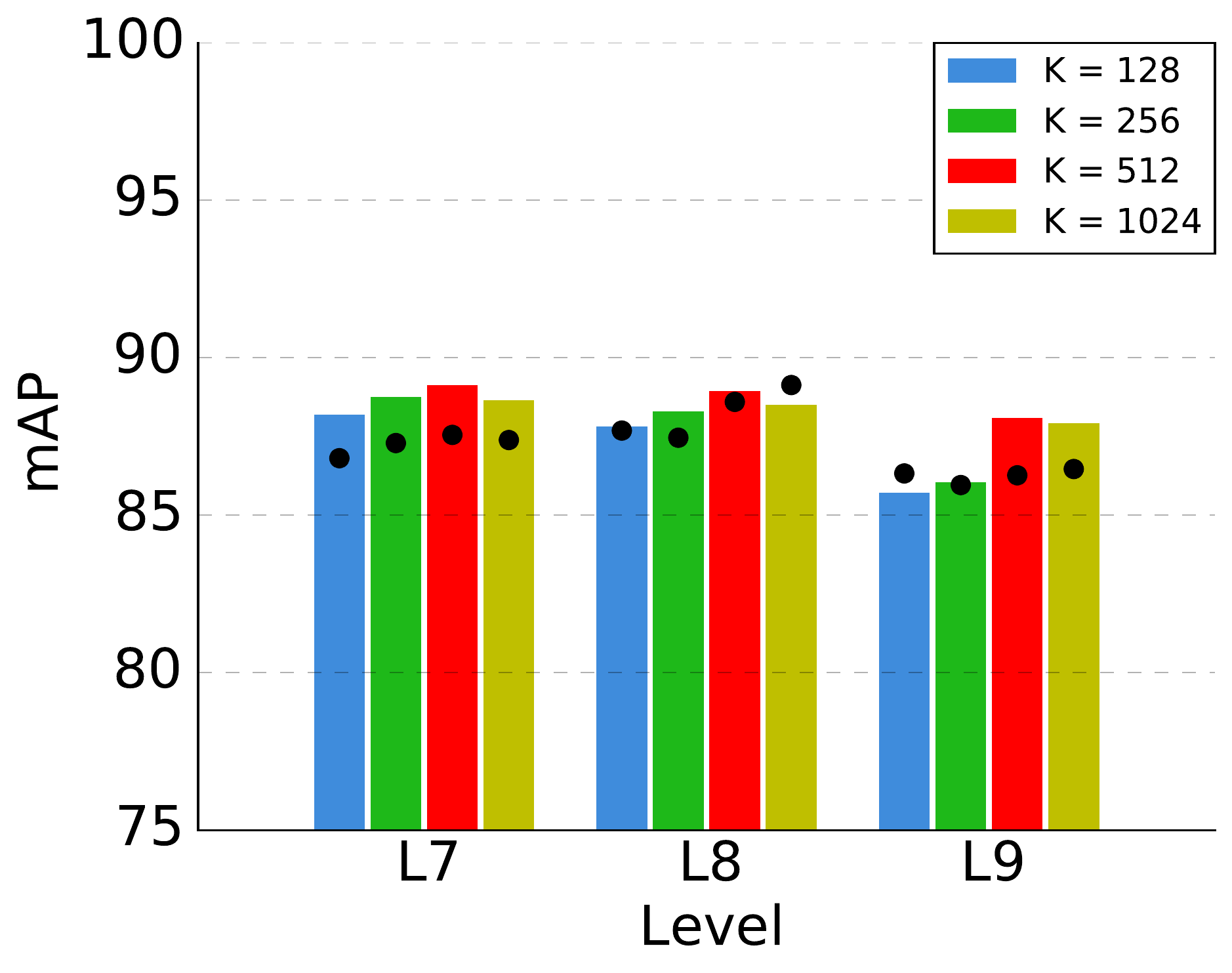}
                 \caption{\small image-to-concept (SVT)}
                 \label{fig2:i2c_svt}
         \end{subfigure}
         \begin{subfigure}[b]{0.24\textwidth}
                 \centering
                 \includegraphics[width=0.85\textwidth]{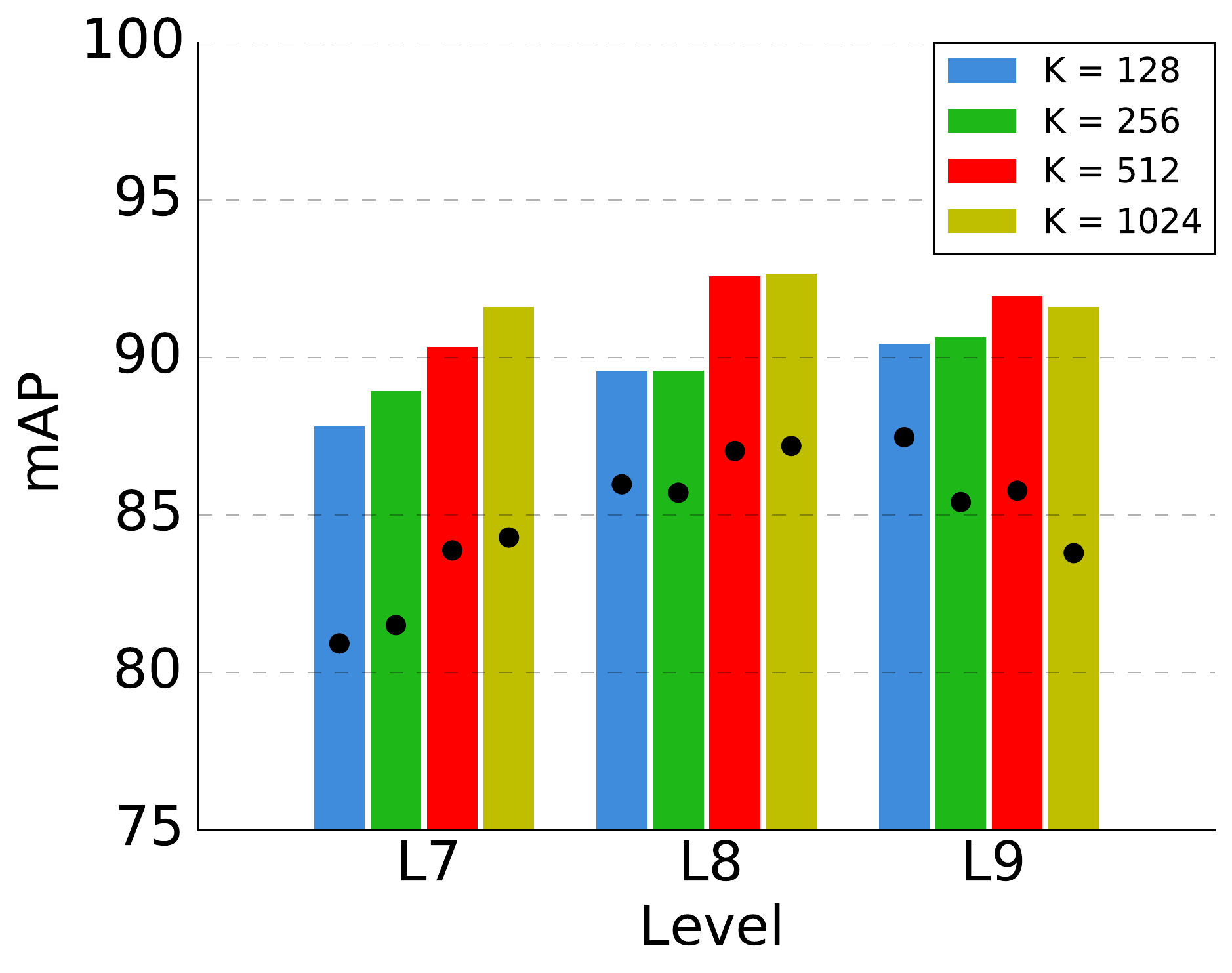}
                 \caption{\small concept-to-image (SVT)}
                 \label{fig2:c2i_svt}
         \end{subfigure}%
         \begin{subfigure}[b]{0.24\textwidth}
                 \centering
                 \includegraphics[width=0.85\textwidth]{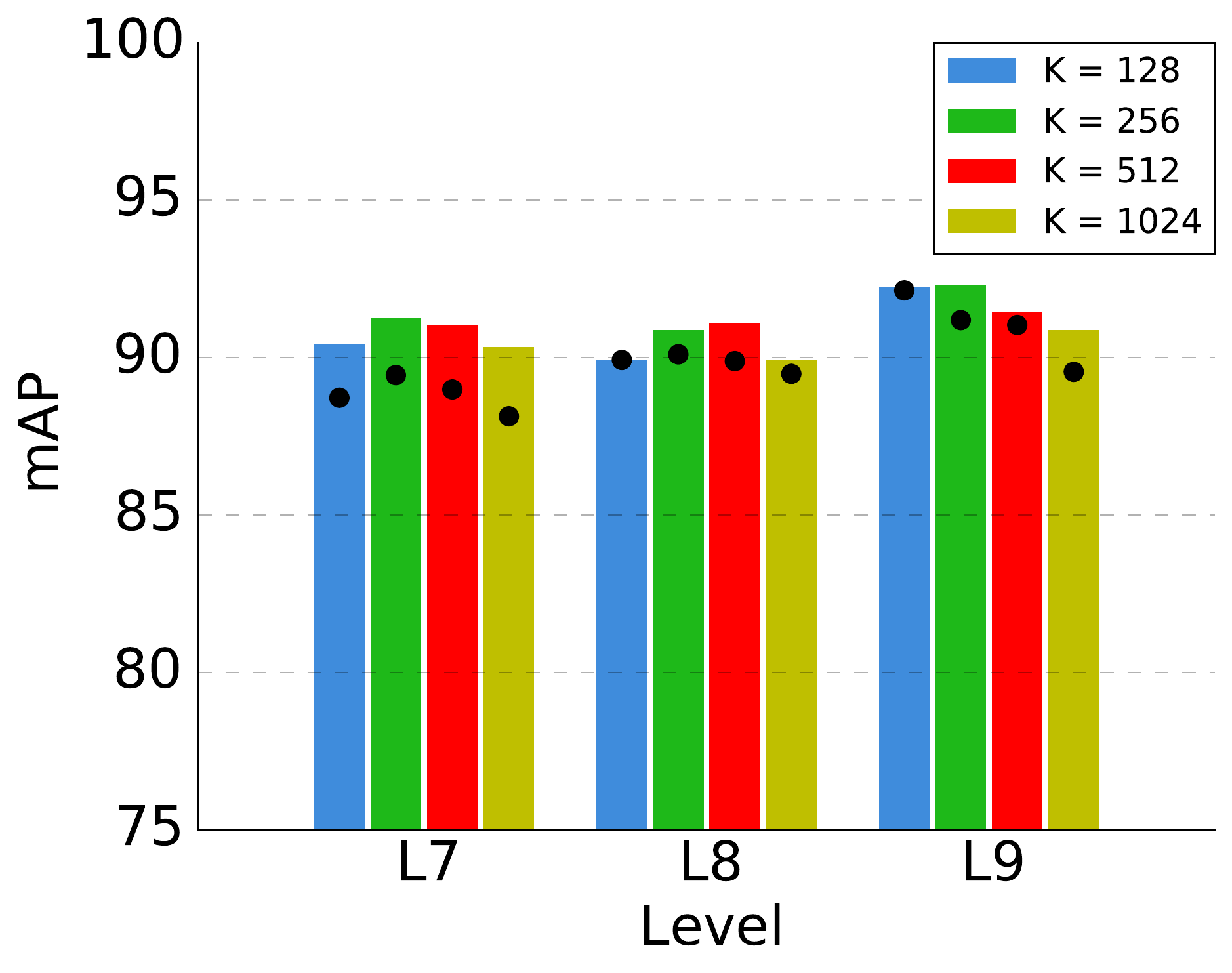}
                 \caption{\small image-to-concept (IIIT5K)}
                 \label{fig2:i2c_iiit5k}
         \end{subfigure}
         \begin{subfigure}[b]{0.24\textwidth}
                 \centering
                 \includegraphics[width=0.85\textwidth]{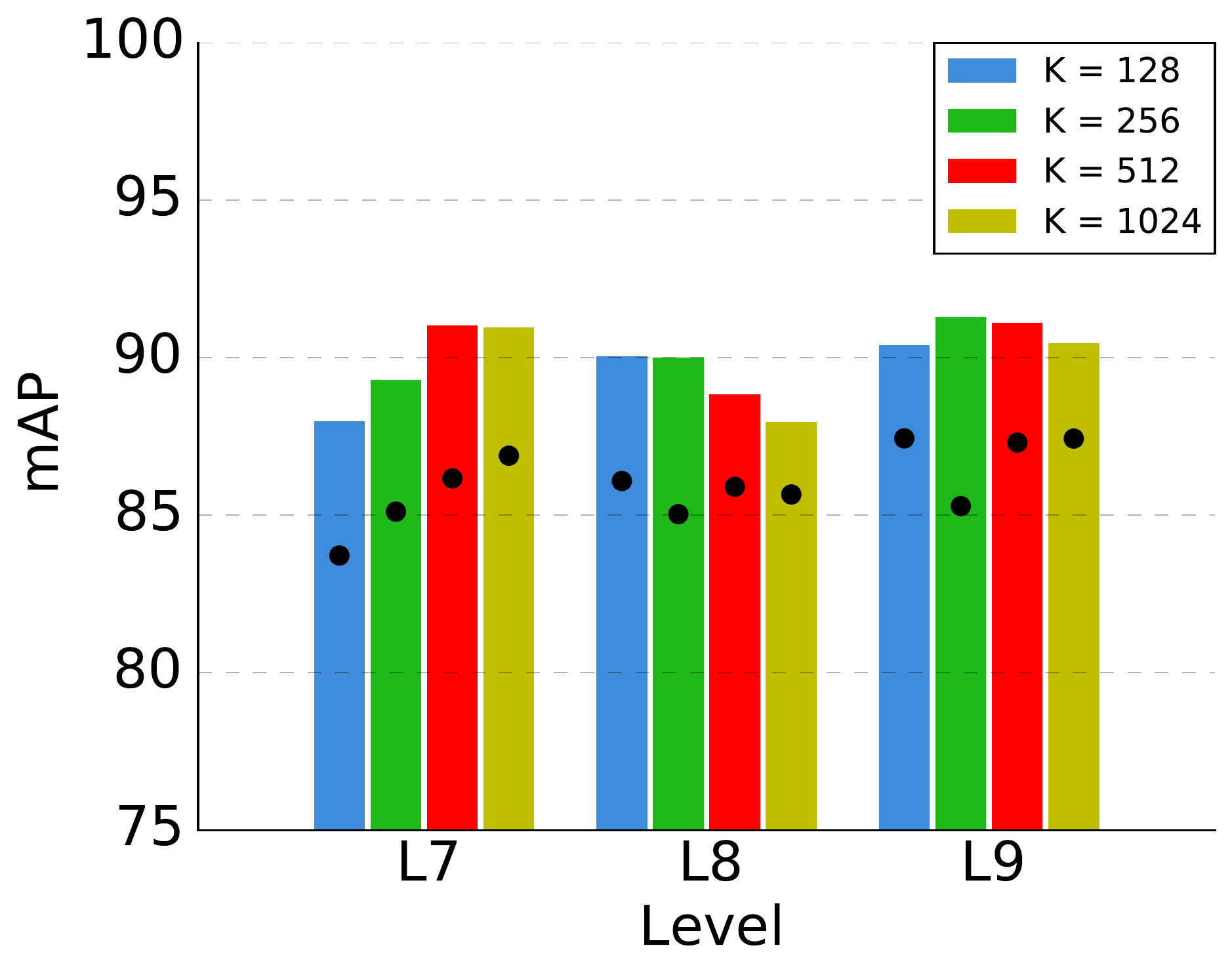}
                 \caption{\small concept-to-image (IIIT5K)}
                 \label{fig2:c2i_iiit5k}
         \end{subfigure}
         \caption{Quantitative results on SVT ((a), (b)) and IIIT5K ((c),(d)). The bars represent the accuracy of the proposed LEWIS method, 
           while the dots represent the accuracy of the two-step baseline.}
        \label{fig:results_svt5k}
\end{figure*}

\subsection{Datasets}

We use three publicly available datasets in our experiments.
The first is the \textbf{Oxford Synthetic Word} dataset \cite{Jaderberg:2014b}, a very large dataset that contains $9$ million annotated word images covering a dictionary of approximately $90{,}000$ English words.
This dataset has been synthetically generated by applying realistic distortions to rendered word images using randomly selected fonts from a catalogue of $1{,}400$ fonts downloaded from Google Fonts.
The official train, validation, and test partitions contain approximately $7.2$ million, $800{,}000$ and $820{,}000$ images, respectively.
Despite being a synthetic dataset, models learned with it obtain outstanding results on real data~\cite{Jaderberg:2014b}.

In addition, we also evaluate the models learned in Oxford Synthetic in two other public datasets: the \textbf{Street View Text (SVT)} dataset \cite{Wang:2011}, which contains a total of $904$ cropped word images harvested from Google Street View, and the \textbf{IIIT 5k-word (IIIT5K)} dataset \cite{Mishra:2012a}, which contains $5{,}000$ cropped word images from natural and born-digital images.
In both cases we only use the official test partitions ($647$ word images in SVT and $3{,}000$ in IIIT5K).

\subsection{Implementation details}
\label{sec:implementation}
To extract the semantic annotations of each word,
we first find the hypernym path to the root for every meaning of the word, as described in Section \ref{sec:wordnet}.
We then keep only the concepts at level $l$ of each path.
In our experiments, we evaluate the effect of varying $l$ from $7$ to $9$.
We follow this approach to extract concepts from the $90{,}000$ words in the Oxford Synthetic dataset.
Concepts are then sorted according to how many words were assigned to them, and only the top $K$ most populated concepts are kept.
In our experiments, we change the value of $K$ from $128$ up to $1{,}024$.
Any word that was not found in the WordNet database or that is not assigned to any concept in the top $K$ is ignored, both at training and at test time.
In the most fine-grained case ($l=9$, $K=128$) this leaves us with $9{,}900$ unique words, and about $820{,}000$ training images and $100{,}000$ testing images.
On the other extreme ($l=7$, $K=$1{,}024) our dataset contains $34{,}153$ unique words, $3{,}000{,}000$ training images, and $350{,}000$ testing images.
The mean number of concepts assigned to every image ranges from $1.2$ and $1.7$, and the maximum number of concepts assigned to a word is $13$.

Our CNN architecture replicates the one in Jaderberg \etal \cite{Jaderberg:2014b}, except for the size of the last layer ($90{,}000$ in their case vs $K$ in our case) and the loss (cross-entropy in their case, and a  WARP ranking loss in our case).
In particular, we use $5$ convolutional layers with ($64$, $128$, $256$, $512$, $512$) kernels of sizes ($5$, $5$, $3$, $3$, $3$) and a stride of $1$ pixel.
A max pooling with size $2$ and a stride of $2$ pixels is applied after layers $1$, $2$, and $4$.
This is followed by three fully connected layers (FC6, FC7, and FC8) of sizes ($4{,}096$, $4{,}096$, $K$).
A ReLU  non-linearity is applied after every convolutional or fully connected layer. Dropout regularization is applied right after layers FC6 and FC7 with a drop rate of $0.5$.
Input images are resized to $32\times 100$ pixels without preserving the aspect ratio, as in \cite{Jaderberg:2014b}.

Learning was done using a modified version of Caffe~\cite{jia2014caffe}.
For efficiency reasons, we first learned $3$ independent models for $l=7$, $l=8$, and $l=9$, fixing the size of $K$ to $128$, and then we fine-tuned those models to larger values of $K$.
Learning all the models took approximately $3$ weeks using $2$ Tesla K40 NVIDIA GPUs.

\subsection{Evaluation protocol}
\vspace{-0.25cm}
We evaluate our approach on three different tasks.
In \textbf{image-to-concept retrieval}, the goal is to annotate a query image with one or multiple concepts. 
This is exactly the task for which our CNN is optimized.
We use each image in the test set of our datasets as a query and use it to retrieve the $K$ concepts ordered by similarity.
The similarity between the word embedding and the concept embeddings is measured as the dot product, and we report mean average precision.
In \textbf{concept-to-image retrieval}, the goal is to retrieve images given a query concept. The similarity between the word embeddings and the concept embedding is also measured as the dot product. In this case, we observed that $\ell_2$-normalizing the image features led to significant improvements. The evaluation metric is also the mean average precision.
In \textbf{image-to-image retrieval}, we are interested in using one image as a query and retrieving other images that share at least one semantic concept.
Images can be represented by the  output of the FC7 layer, which corresponds to the latent space, but also by the output of the last layer, which would correspond to an ``attribute scores'' layer, where the image is represented by stacking the similarities between the image and all $K$ concepts.
This is a more challenging task, since two images that have many different associated concepts but share one of them are still considered a match. In this case, we report precision at $k$, for values of $k$ of $1$, $10$, and $50$, and R-Precision, where the number of relevant images per query is used as cutoff.

We consider  \textbf{two baselines} in our experiments.
The first one is the two-step approach based on transcribing the word image and matching the transcriptions. 
For this task we use a state-of-the-art dictionary CNN \cite{Jaderberg:2014b}.
We use the pretrained model that the authors made available. 
This model achieves around $95\%$ transcription accuracy on the Oxford Synthetic dataset by choosing the right transcription out of a pool of $90{,}000$ candidates.
In this baseline, we first use this model to choose the most likely transcription of a given image, and then we propagate concepts extracted from WordNet using that transcription.
This allows us to match an image with concepts, and to perform both image-to-concept and query-by-image retrieval using inverted indices.

As a second baseline, we use the output activations of the penultimate (FC7) layer of the same model as a feature representation of the words ($4{,}096$ dimensions).
This is a very strong feature representation that encodes information about the characters of the word. 
We denote it with CNN-Dict$_7$.
These features can subsequently be used for image-to-image retrieval, or for concept prediction after learning a linear classifier on top.

We also evaluate the effect of inaccurate cropping of the word images.
In most realistic scenarios involving end-to-end tasks, it is necessary to localize and crop the word images out of larger images.
Even if the localization techniques have improved in recent years, localization is still inexact at best.
To test the effect of this, as a surrogate of text localization, we perform random crops of the word images, randomly removing up to $20\%$ of the image from left and right and up to $20\%$ of the image from top and bottom.
All of these cropped images still have an intersection over union with the originals larger or equal than $(1-0.2)^2=0.64$, 
and would be accepted as positive localizations using the standard localization threshold of $0.5$.

\begin{table}
    \footnotesize
    \centering
    \caption{Image-to-image retrieval with $K=256$ concepts. We compare our features with the CNN-Dict$_7$ features \cite{Jaderberg:2014b}.}
    \begin{tabular}{@{}l|l|cccc@{}}
                            &      & P@1 & P@10 & P@50 & R-P \\ \midrule
        \multirow{3}{*}{l7}& CNN-Dict$_7$  & \textbf{96.20} &83.99 &29.67 &4.04 \\
                           & LEWIS (FC7)  &  95.10 &\textbf{91.89} &49.08 &12.21 \\
                           & LEWIS (FC8)  & 94.16 &91.79 &\textbf{61.73} &\textbf{29.05} \\
        \midrule
        \multirow{3}{*}{l8} & CNN-Dict$_7$  & \textbf{96.33} &84.45 &29.74 &4.20\\
                            & LEWIS (FC7)   & 95.50 &92.43 &49.73 &13.21 \\
                            & LEWIS (FC8)   & 94.67 &\textbf{92.68} &\textbf{67.99} &\textbf{37.79} \\
        \midrule
        \multirow{3}{*}{l9}        & CNN-Dict$_7$   &  \textbf{96.64} &85.49 &30.30 &4.90 \\
                                   & LEWIS (FC7)  &  95.80 &93.01 &49.20 &13.64 \\
                                   & LEWIS (Last)  & 95.00 &\textbf{93.28} &\textbf{68.58} &\textbf{39.34} \\
    \end{tabular}
    \label{tab:im2im}
    \vspace{-0.25cm}    
\end{table}

\begin{figure*}
\centering
\includegraphics[width=0.77\linewidth]{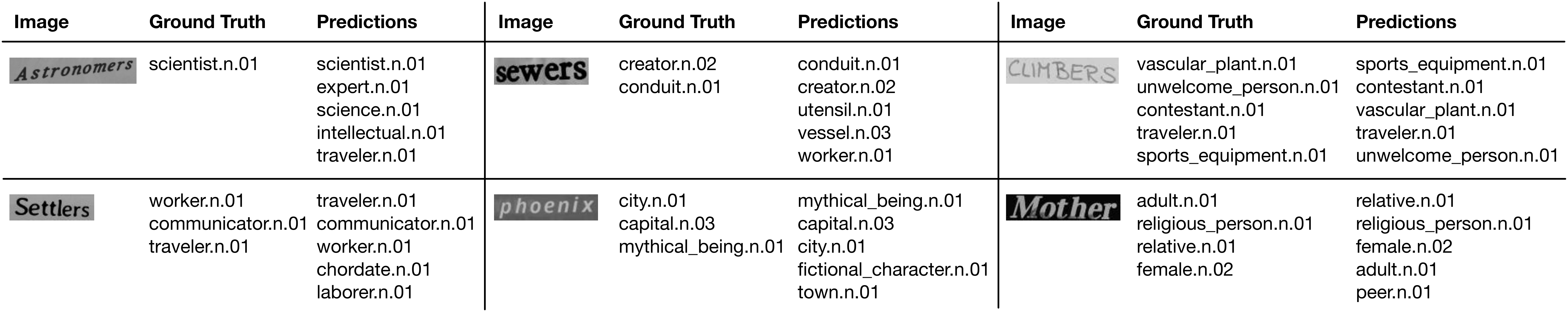}
\caption{Qualitative results on the image-to-concept task with $K=128$ and concepts from levels $7$ and $8$. Many of the top predicted concepts are meaningful even if they do not appear amongst the ground truth ones.}
\label{fig:qres_i2c}
\end{figure*}

\begin{figure*}
\centering
\includegraphics[width=0.75\linewidth]{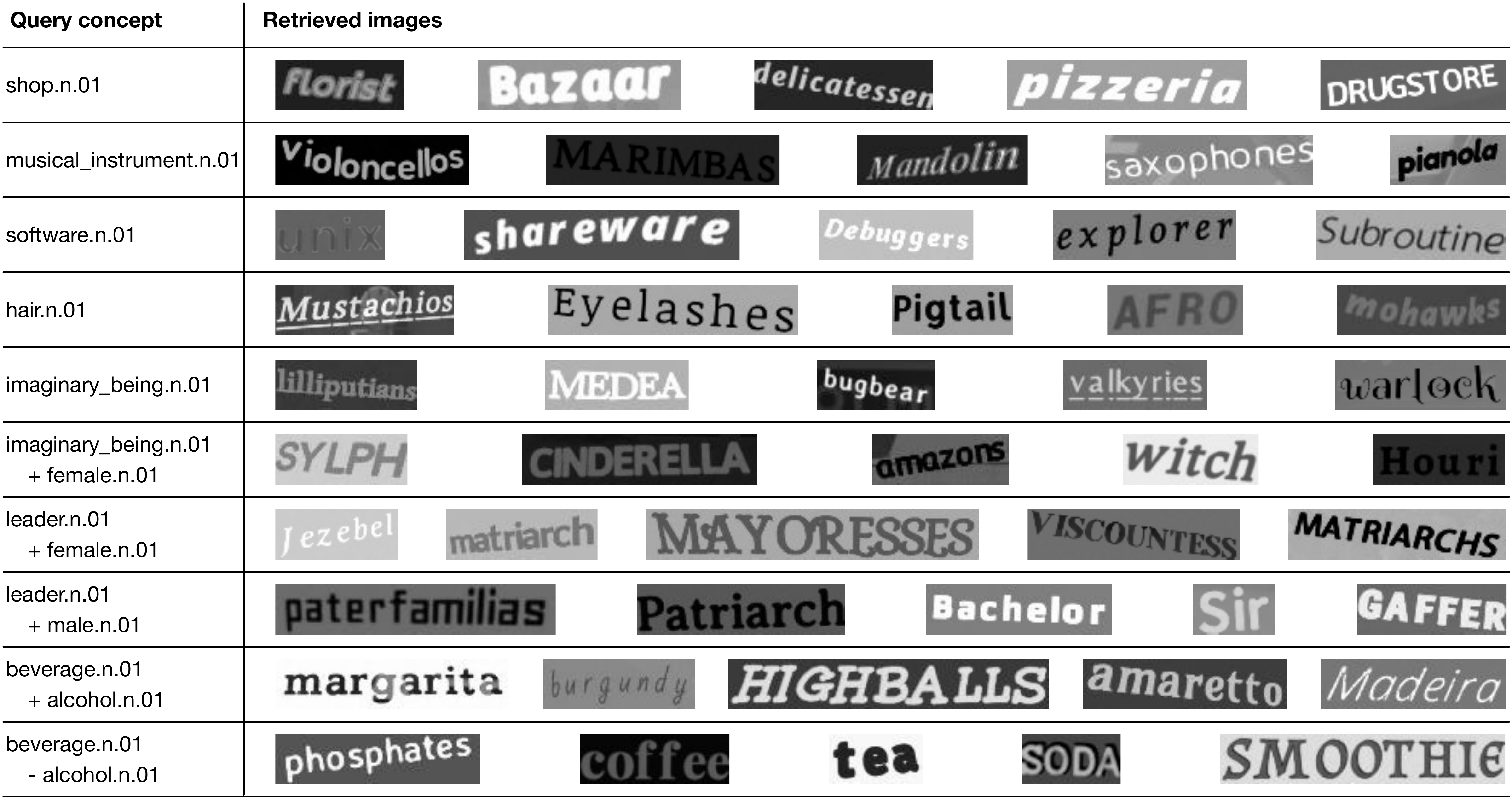}
\caption{Qualitative results on the concept-to-image task with $K=128$ and concepts from levels $7$ and $8$. For every concept, we show images of unique words returned in the first positions. No negative image was ranked ahead of any of these images.}
\label{fig:qres_c2i}
\end{figure*}

\subsection{Results and discussion}
\vspace{-0.15cm}
\paragraph{Image-to-concept and concept-to-image retrieval tasks}
We first evaluate the proposed LEWIS approach on the image-to-concept and concept-to-image tasks on Oxford Synthetic and compare it with the two-step baseline.
We report results in Figure \ref{fig:results_oxford}, (a) and (b).
The bars represent our approach, and the dots denote the two-step baseline.

We observe that increasing the number of levels (\ie, more fine-grained concepts) generally leads to improved results.
This is reasonable, as in the extreme case of one concept per word this is equivalent to the transcription problem, which we know can be addressed with a CNN.
On the other hand, less depth implies more words per semantic concept.
This leads to a more multimodal problem, where very different images have to be assigned to the same class, increasing the difficulty.
Increasing the number of concepts $K$ also has a limited impact.
As the concepts were sorted by number of words assigned, the first concepts are more difficult than the subsequent ones, leading to a trade-off between having more concepts, but these concepts being easier.

Compared to the two-step baseline, our method is slightly behind (about $1$ percent absolute) on the image-to-concept task. However, in the concept-to-image task, we outperform the baseline. We believe the concept-to-image task is less forgiving towards badly transcribed images, as they affect negatively several queries, and that the softer nature of the proposed embeddings makes them more suitable for the task.
We also evaluate on images with random crops in (c) and (d). In this case, as expected, the recognition baseline fails, while our approach is still able to detect key aspects of the word image and favor the appropriate concepts.
We evaluate as well on the SVT and IIIT5K datasets (Figure \ref{fig:results_svt5k}), where the results exhibit a similar behaviour.
Despite having learned the models on Oxford Synthetic, the results on SVT and IIIT5K are still very accurate.

We believe that learning the features with a deep architecture focused on the semantics is a key factor in our approach.
This is demonstrated through the comparison with the second baseline:
we use the state-of-the-art CNN-Dict$_7$ features \cite{Jaderberg:2014b} and learn a linear classifier 
using the same ranking loss and data we use to train LEWIS for a fair comparison.
In this case, we only achieve an accuracy of $59\%$ mean average precision on the image-to-concept task ($K=128$, $l=7$), 
compared to the $95\%$ achieved when learning the features with the semantic goal in mind.
This shows that, to perform these types of tasks, traditional or recent word image features 
that only encode character information are not suitable, and that it is necessary to learn and encode these semantics directly in the representation.

\vspace{-0.5cm}
\paragraph{Image-to-image retrieval}
We now focus on the image-to-image task, where one image is used as a query and the goal is to return all the images that are related, \ie, that have at least one concept in common at a given level.
We compare the proposed LEWIS features, extracted from the previous-to-last layer (FC7, $4{,}096$ dimensions) and the last layer (FC8, $K$ dimensions), with the CNN-Dict$_7$ features of Jaderberg et al \cite{Jaderberg:2014b}. 
All features are $\ell_2$-normalized and compared using the dot product.

Table \ref{tab:im2im} reports results for $K=256$ at several levels using precision @$1$, @$10$, @$50$, and R-Precision as metrics.
At precision @$1$, the CNN-Dict$_7$ features obtain superior results, as they are returning another image with exactly the same word and they do this with better accuracy.
However, as $k$ increases and images with different words need to be returned, its accuracy plummets, as this representation only encodes information to recognize the exact word.
On the other hand, our embeddings still return meaningful results when $k$ increases, even if they have not been learned explicitly for this type of retrieval task.

\vspace{-0.5cm}
\paragraph{Qualitative results}
Figure \ref{fig:qres_i2c} illustrates some qualitative results for the image-to-concept task, using $K=128$ classes and depth levels $7$ or $8$.
In many cases, the predicted concepts are very related to the query even if they do not appear in the ground truth annotations, showing that semantically similar concepts are being embedded in neighboring locations of the space.
Figure \ref{fig:qres_c2i} shows qualitative results for the concept-to-image tasks, showing once again that images with very different transcriptions are still embedded close to their related concepts.
Interestingly, we can combine concepts, by adding or subtracting the scores, to make more complex searches that still return meaningful results.

\vspace{-0.5cm}
\paragraph{Generalization}
One of the advantages of our method with respect to the baseline is that we can encode and find concepts 
for words that have not been observed during training or that do not appear in WordNet.
The previous experimental results hinted that some generalization has been achieved.
For example, the qualitative results of Figure \ref{fig:qres_i2c} showed that some concepts were predicted based on the roots of similar words, 
as those concepts did not appear in the ground truth of the words.
This is consistent with the results using random crops, where reasonable results were obtained even if part of the word was missing.
Here we test this explicitly by training a network on a subset of the training data ($90\%$ of words) 
and testing on a disjoint set ($10\%$ of words), where none of the testing words were observed during training.
In this case, the results dropped from around $90\%$ down to $56.1\%$ ($K=128$, $l=7$) and $61.9\%$ ($K=256$, $l=7$) in image-to-concept task, 
and to $40.6\%$ and $52.8\%$ in concept-to-image.
Although there is a significant drop in accuracy, the results are still surprisingly high, 
given that this is a very arduous zero-shot problem.
This shows that some generalization to new words is indeed achieved, likely through common roots of words.

\section{Conclusions}
\label{sec:conclusions}
In this paper we have introduced a new task to the computer vision community: predicting relevant semantic categories for word images.
We believe that solutions to this task can greatly benefit problems related to scene text, particularly urban scene understanding.
To address this new task, we propose an approach based on CNNs that learns to rank semantic concepts in an end-to-end manner, starting directly from the image pixels.
The proposed approach, LEWIS, can be understood as learning an embedding space shared by both word images and semantic concepts.
LEWIS performs similarly to or outperforms a two-step baseline based on a state-of-the-art word transcription method on a variety of tasks, while offering significant advantages.
We also generated semantic annotations for an existing large-scale word image database, which we will share with the community to help further research on this task.

{\small
\bibliographystyle{ieee}
\bibliography{egbib}
}

\end{document}